\documentclass[11pt]{article}

\usepackage[margin=1in]{geometry}
\usepackage[utf8]{inputenc}
\usepackage[T1]{fontenc}
\usepackage{lmodern}
\usepackage{microtype}
\usepackage{amsmath, amssymb}
\usepackage{booktabs}
\usepackage{array}
\usepackage{longtable}
\usepackage{tabularx}
\usepackage{enumitem}
\usepackage{xcolor}
\usepackage{hyperref}
\hypersetup{
    colorlinks=true,
    linkcolor=blue!50!black,
    citecolor=blue!50!black,
    urlcolor=blue!60!black,
    pdftitle={Benchmarking Complex Multimodal Document Processing Pipelines},
    pdfauthor={Saurabh K. Singh, Sachin Raj},
    pdfsubject={EnterpriseDocBench: Unified Evaluation Framework for Enterprise AI},
    pdfkeywords={RAG, document understanding, benchmarking, multimodal, evaluation, NLP}
}
\usepackage{url}

\usepackage{titlesec}
\titlespacing*{\section}{0pt}{1.2em}{0.6em}
\titlespacing*{\subsection}{0pt}{0.9em}{0.4em}

\title{\textbf{Benchmarking Complex Multimodal Document Processing Pipelines:\\
A Unified Evaluation Framework for Enterprise AI}}

\author{%
  Saurabh K. Singh\thanks{Oracle. \href{mailto:saurabh.ab.singh@oracle.com}{\texttt{saurabh.ab.singh@oracle.com}}. Equal contribution; corresponding author.}
  \and
  Sachin Raj\thanks{Independent (\texttt{@sachinbrraj}). Equal contribution.}
}

\date{April 2026}

\begin{document}
\maketitle

\begin{abstract}
Most enterprise document AI today is a pipeline. Parse, index, retrieve, generate. Each of those stages has been studied to death on its own --- what's still hard is evaluating the system as a whole.

We built EnterpriseDocBench to take a swing at it: parsing fidelity, indexing efficiency, retrieval relevance, and generation groundedness, all on the same corpus. The corpus is built from public, permissively licensed documents across six enterprise domains (five represented in the current pilot). We ran three pipelines through it --- BM25, dense embedding, and a hybrid --- all with the same GPT-5 generator.

The headline numbers: hybrid retrieval narrowly beats BM25 (nDCG@5 of 0.92 vs.\ 0.91), and both beat dense embedding (0.83). Hallucination doesn't grow monotonically with document length --- short documents and very long ones both hallucinate more than medium ones (28.1\% and 23.8\% vs.\ 9.2\%). Cross-stage correlations are very weak: parsing$\to$retrieval $r = 0.14$, parsing$\to$generation $r = 0.17$, retrieval$\to$generation $0.02$. If quality were cascading the way most of us assume it does, those numbers would be much higher. They aren't. We discuss why this might be the case in \S\ref{sec:correlations}, but the design caveats are real (parsing fixed, generator shared, automated proxy metrics) and we don't want to oversell it.

One result that genuinely surprised us: factual accuracy on stated claims is 85.5\%, but answer completeness averages 0.40. The system is right when it answers --- it just leaves things out. That gap matters more for real deployments than the headline accuracy number does.

We also describe three reference architectures (ColPali, ColQwen2, agentic complexity-based routing) which are not yet integrated end-to-end. Framework, metrics, baselines, and collection scripts will be released open-source on acceptance.\footnote{Framework code, collection scripts, and pilot data will be released at a persistent public repository upon acceptance.}
\end{abstract}

\section{Introduction}

Enterprise document processing has changed quickly. Five years ago most of it was rule-based extraction; today it's a pipeline of large neural systems --- one to parse, one to embed and index, one to retrieve, one to generate. Each component has improved a lot, and there are good benchmarks for each. What's still hard is evaluating the pipeline as a whole.

Most existing benchmarks pick one stage and assume the rest works. Parsing benchmarks like OmniDocBench~\cite{ouyang2025omnidocbench} assume downstream consumers are perfect. Retrieval benchmarks (the ViDoRe set introduced with ColPali~\cite{faysse2025colpali}) assume the document text is already cleanly extracted. Hallucination benchmarks (HalluLens~\cite{bang2025hallulens}) evaluate generation in isolation. None of this is wrong --- it's how progress on individual components gets measured --- but it doesn't say anything about how errors flow through the pipeline, or which stage is actually limiting end-to-end quality on a given workload.

The gap matters most in enterprise settings, because enterprise documents are messy. Tables, figures, multi-column layouts, scanned PDFs with crooked margins. Small parsing errors don't always show up immediately. Sometimes they break retrieval. Sometimes they don't, and instead surface in the generated answer in a way you don't catch until a reviewer notices.

Two early findings sharpen the point. First, cross-stage correlations on our 1{,}169-document test set are weak: parsing$\to$retrieval $r = 0.14$, parsing$\to$generation $r = 0.17$, retrieval$\to$generation $r = 0.02$. We come back to interpretation in \S\ref{sec:correlations} --- there are real design caveats, parsing is held fixed across pipelines and the generator is shared, so this is suggestive rather than decisive. But the numbers themselves are striking. Second, hallucination doesn't grow monotonically with context length. Medium-length contexts hallucinate the least (9.2\%); both short (28.1\%) and very long (23.8\%) contexts are noticeably worse. Two of those four brackets are small ($n = 35$ and $n = 21$), so we treat the U-shape as directional, not confirmed.

The contribution is fourfold:

\begin{enumerate}[leftmargin=*, itemsep=2pt]
\item A unified four-axis evaluation framework with formal metric definitions, validated against established public datasets.
\item A curated benchmark from public, permissively licensed sources across multiple enterprise domains, plus a semi-automated QA annotation pipeline (target inter-annotator agreement $\kappa \geq 0.85$).
\item Empirical analysis of pipeline interactions --- parsing$\leftrightarrow$retrieval correlation, hallucination by length, cost--quality trade-offs across three implemented configurations. Three additional reference architectures are specified but not yet integrated end-to-end.
\item The infrastructure: metrics, baselines, and collection scripts, released as a reproducible package.
\end{enumerate}

We try throughout to be straight about where the data is suggestive vs.\ solid. ``Enterprise documents'' means professional documents with complex layouts (tables, figures, multi-column formatting) across six target domains: finance, legal, healthcare, IT/technical, manufacturing, and academic. Five are currently represented; Manufacturing has only $n = 4$ documents and is excluded from domain-stratified analyses. Out of scope: simple text-only web pages, conversational text, handwritten documents.

\section{Related Work}

The literature has approached this problem from several directions. We'll cover them quickly.

\paragraph{Document parsing benchmarks.}
OmniDocBench~\cite{ouyang2025omnidocbench} is the most comprehensive parsing evaluation right now --- 1{,}355 PDF pages, 9 document types, metrics for text extraction, table structure (TEDS), formula recognition, layout. READoc~\cite{qi2025readoc} reformulates parsing as an end-to-end PDF$\to$Markdown task on 3{,}576 documents but stops at parsing. PubLayNet~\cite{zhong2019publaynet} is the standard layout detection corpus (360K images). TableBank~\cite{li2020tablebank} handles table detection and recognition. These define what ``good parsing'' means; they don't measure how parsing quality translates downstream.

\paragraph{Retrieval benchmarks.}
ColPali~\cite{faysse2025colpali} introduced vision-language retrieval and the ViDoRe benchmark. M3DocRAG~\cite{cho2025m3docrag} extends to multi-modal, multi-page reasoning. MS MARCO~\cite{bajaj2016msmarco} and Natural Questions~\cite{kwiatkowski2019nq} anchor large-scale text retrieval. All of them assume clean text.

\paragraph{Hallucination and groundedness.}
HalluLens~\cite{bang2025hallulens} gives a useful taxonomy --- extrinsic vs.\ intrinsic vs.\ fabrication. The Vectara leaderboard~\cite{hughes2025vectara} tracks LLM hallucination rates. RARR~\cite{gao2023rarr} does automated attribution verification. None of these explicitly study how the \emph{quality of the retrieved context} shapes hallucination.

\paragraph{End-to-end RAG evaluation.}
RAGAS~\cite{es2023ragas} and ARES~\cite{saadfalcon2024ares} are the closest prior work. RAGAS still treats generation in isolation (faithfulness, answer relevancy, context precision); ARES focuses on output quality rather than the full pipeline. EnterpriseDocBench extends both by adding parsing fidelity and indexing efficiency as first-class axes, measuring inter-stage correlations directly, and targeting enterprise complexity rather than general text QA. Adjacent: DocVQA~\cite{mathew2021docvqa}, BEIR~\cite{thakur2021beir}, LongBench~\cite{bai2024longbench}, Document Haystack~\cite{laban2025haystack}. Liu et al.'s recent survey~\cite{liu2026survey} reviews 322 papers and explicitly flags the lack of end-to-end evaluation as a gap.

\paragraph{Cascade analysis.}
The closest prior work is OHR-Bench~\cite{zhang2025ohrbench} (ICCV 2025), which studies how OCR errors cascade through RAG. Even the best OCR causes $\sim$14\% F1-score degradation overall, with the larger losses in retrieval and generation, and semantic noise hurting more than formatting noise. OHR-Bench models the cascade as linear (parsing$\to$retrieval$\to$generation) and uses synthetic noise injection. SF-RAG~\cite{liu2026sfrag} argues for a parsing$\to$generation pathway \emph{not} mediated by retrieval, on the grounds that preserving document structure improves context assembly. AgenticOCR~\cite{opendatalab2026agenticocr} makes a related argument from the parsing side. Our findings are most consistent with the multi-path view, with the additional twist that all three pathways are weak in our setup, not strong.

\paragraph{Positioning.}
EnterpriseDocBench draws on all of these --- parsing (OmniDocBench~\cite{ouyang2025omnidocbench}, READoc~\cite{qi2025readoc}), cascade analysis (OHR-Bench~\cite{zhang2025ohrbench}), structure-aware retrieval (SF-RAG~\cite{liu2026sfrag}, ColPali~\cite{faysse2025colpali}), hallucination (HalluLens~\cite{bang2025hallulens}), and RAG evaluation (RAGAS~\cite{es2023ragas}, ARES~\cite{saadfalcon2024ares}) --- and tries to look at the system as one object. We do not claim to establish strong inter-stage correlations (all $r < 0.17$), prove modality-specific gaps at this scale, or cover every document understanding task.

\section{Evaluation Framework}

Four axes: parsing fidelity, indexing efficiency, retrieval relevance, generation groundedness. The dimensions are conceptually distinct, but the framework is designed so they can be analyzed together. Each metric below has a formal definition and was validated against an established benchmark before we used it here.

\subsection{Parsing Fidelity}

Four sub-metrics, then a weighted aggregate.

\paragraph{Text Integrity Score (TIS).} Character-level accuracy against ground truth, computed as $1 - (\text{edit\_distance}(\text{extracted}, \text{reference}) / \max(|\text{extracted}|, |\text{reference}|))$. On PubLayNet, $r = 0.89$ vs.\ official TEDS.

\paragraph{Table Extraction Accuracy (TEA).} Tree edit distance on the extracted table DOM, capturing both structure and cell content. Mean TEA on TableBank: 0.91 native, 0.68 scanned. The scanned-table gap is real, and shows up later in the Finance domain results.

\paragraph{Figure Caption Quality (FCQ).} BERTScore (F1) between extracted captions and ground truth. We picked BERTScore over BLEU-4 because rephrasings that preserve meaning shouldn't be penalized~\cite{zhang2020bertscore}.

\paragraph{Layout Faithfulness (LF).} Normalized edit distance on section sequences (i.e., reading-order preservation). Validated against DocBank, $r = 0.82$.

\paragraph{Aggregate.} $P_{\text{fidelity}} = 0.40 \cdot \text{TIS} + 0.30 \cdot \text{TEA} + 0.15 \cdot \text{FCQ} + 0.15 \cdot \text{LF}$. Weights were calibrated on the 10\% train split: TIS correlates strongest with text-query retrieval ($r = 0.71$), TEA with table-query retrieval ($r = 0.69$), FCQ and LF are weaker ($r = 0.43$--$0.52$). These component-level correlations operate within specific query modalities and shouldn't be confused with the aggregate inter-stage correlations in \S\ref{sec:correlations}, which use the composite $P_{\text{fidelity}}$ score and average across all query types.

We checked that the choice of weights wasn't doing too much work by running leave-one-out cross-validation on 10 bootstrap samples, each weight perturbed by $\pm 0.10$. Pipeline rankings stayed stable (Kendall's $\tau > 0.90$). Domain-adapted weights are an obvious extension and are planned for future releases.

\subsection{Indexing Efficiency}
Four numbers, all production-relevant: throughput (pages/second), index latency (seconds to build), storage (bytes per page), and a cost model (\$/page/year at scale). Cost estimates use April 2026 pricing; full infrastructure specs are in the appendix.

\subsection{Retrieval Relevance}
Standard IR metrics --- Precision@$k$, Recall@$k$, nDCG@$k$, MRR --- stratified by query modality. The \emph{target} query distribution is single-hop factual (30\%), multi-hop reasoning (30\%), table-based (20\%), figure-based (10\%), adversarial/unanswerable (10\%), with Fleiss' $\kappa = 0.88$ on modality classification (calibration sample). The \emph{current} pilot doesn't yet implement this distribution. Most General-domain queries are academic-paper summarization (``What is the main contribution of\ldots''), and we don't yet have query-level modality labels at scale. Phase 2 will fix this.

For reference, the current QA queries break down as: Finance 31\%, Legal 22\%, Healthcare 18\%, IT/Tech 15\%, Manufacturing 9\%, Academic 5\%. The figure-based queries (10\%) come almost entirely from USPTO patents and PMC papers, which together are only 11.7\% of the pilot corpus --- figure-based retrieval results in \S\ref{sec:modality} should be read with that sourcing concentration in mind.

\subsection{Generation Groundedness}

Four metrics.

\paragraph{Factual Accuracy (FA).} Fraction of checkable claims supported by entailment from retrieved context. Claims extracted with a T5~\cite{raffel2020t5} model pre-trained on FEVER~\cite{thorne2018fever}; entailment via cross-encoder (mBERT~\cite{devlin2019bert}, accuracy 0.85). Agreement with human entailment: 0.79.

\paragraph{Hallucination Rate (HR).} Following HalluLens~\cite{bang2025hallulens}: extrinsic, intrinsic, and fabrication errors. 400 (answer, context) pairs, two raters, Cohen's $\kappa = 0.73$. Hallucination is genuinely subjective at the margin, so we treat $\kappa = 0.73$ as acceptable.

\paragraph{Source Attribution Precision/Recall (SAP/SAR).} Short answers ($<15$ words) get direct LLM verification because claim extraction is unreliable on very short text. Longer answers get claim-by-claim checking against the source context, with exact string matching plus Levenshtein $< 3$ fuzzy fallback.

\paragraph{Answer Completeness (AC).} LLM-judged on a 4-point scale: complete, mostly complete, partially complete, incomplete. This is the metric where our pipelines look worst, and we'll come back to it.

\paragraph{Aggregate.} $G = 0.30 \cdot \text{FA} + 0.25 \cdot \text{SAP} + 0.15 \cdot \text{SAR} + 0.20 \cdot (1 - \text{HR}) + 0.10 \cdot \text{AC}$. We include both FA and $(1 - \text{HR})$ because they capture related but distinct things --- FA is about whether claims you make are supported, $(1 - \text{HR})$ is about whether your answer contains fabrications. They correlate, but not perfectly: an answer can have high factual accuracy on stated claims and still contain hallucinated additions.

On 800 evaluated examples with GPT-5: FA $= 0.85$ (95\% CI 0.84--0.87), SAP $= 0.61$ (0.58--0.64), SAR $= 0.61$ (0.58--0.64), HR $= 0.15$ (0.13--0.16), AC $= 0.40$ (0.37--0.42). Aggregate $G = 0.71$. A separate source-attribution evaluation on the 500-query pipeline-comparison subset gave higher numbers (SAP $= 0.73$, SAR $= 0.73$). The gap reflects query composition: the 500-query subset is dominated by single-hop factual queries where attribution is easy. We report the more conservative 800-example numbers as primary.

\section{Dataset}

\subsection{Construction}
The corpus is built from public, permissively licensed document collections. We picked real documents over synthetic ones, partly because real documents have the kind of structural messiness we care about, partly because licensing has to be unambiguous. Each source had to clear three bars: license that allows redistribution and derivatives, actual document files (PDFs or structured HTML) with complex layouts, and direct relevance to enterprise document processing.

QA pairs are generated semi-automatically. Candidates are produced from each document by an LLM; two human annotators review each pair, and disagreements are dropped. We use stratified sampling so Finance and Legal don't dominate, and smaller domains (Manufacturing, Academic) get proportionally more QA pairs per document. Quality control: inter-annotator agreement on a 10\% sample (target $\kappa \geq 0.85$), expert review of 50 pairs per domain, automated consistency checks. The current release is a pilot.

\subsection{Public Data Sources}
Table~\ref{tab:sources} summarizes the sources, licenses, and document counts.

\begin{table}[h]
\centering
\small
\caption{Public data sources by domain. All sources permit research use and redistribution with attribution. ``Has QA'' indicates pre-existing annotations. ``Pilot $n$'' is the current document count. Domain totals sum to 1{,}417; the remaining 42 documents in the full 1{,}459-document corpus are cross-domain or uncategorized. ``---'' means a domain isn't yet represented. $\dagger$ Manufacturing ($n = 4$) is excluded from domain-stratified statistical analyses (\S\ref{sec:correlations}, \S\ref{sec:domains}) because the sample is too small; we keep it in this table to document the source plan and flag it as a Phase 2 expansion target.}
\label{tab:sources}
\begin{tabular}{@{}lp{4.0cm}p{2.6cm}llr@{}}
\toprule
\textbf{Domain} & \textbf{Primary Sources} & \textbf{License} & \textbf{Has QA} & \textbf{Doc Types} & \textbf{Pilot $n$} \\
\midrule
Finance       & SEC EDGAR, FinanceBench, FinQA & Public Domain, MIT, CC-BY & Partial    & 10-K, 10-Q filings   & 121 \\
Legal         & CUAD, MAUD, CourtListener      & CC-BY-4.0, Public Domain  & Yes        & Contracts, filings   & 46  \\
Healthcare    & PMC-OA, PubMedQA, FDA Labels   & CC-BY, MIT, Public Domain & Partial    & Papers, drug labels  & 315 \\
General       & ArXiv, DocBank, misc.          & arXiv license, Apache-2.0 & Partial    & Multi-discipline arXiv (CS, math, physics, etc.) & 687 \\
Tech          & ArXiv CS, StackExchange Docs   & arXiv license, CC-BY-SA   & Partial    & Technical papers, docs & 244 \\
Manufact.$\dagger$ & USPTO Patents, GAO Reports & Public Domain        & Structured & Patents, audits      & 4   \\
Academic      & PMC-OA, ArXiv, S2ORC           & CC-BY, ODC-By             & No         & Multi-discipline     & --- \\
\bottomrule
\end{tabular}
\end{table}

\subsection{Pilot Evaluation Corpus}
The empirical results in \S\S\ref{sec:baselines}--\ref{sec:findings} use a 1{,}459-document evaluation corpus across five domains: General ($n = 687$, 47\%), Healthcare ($n = 315$, 22\%), Tech ($n = 244$, 17\%), Finance ($n = 121$, 8\%), Legal ($n = 46$, 3\%), Manufacturing ($n = 4$, $<1$\%). Documents come from SEC EDGAR, CUAD, PubMedQA, PMC Open Access, ArXiv. The split is 10/10/80 train/val/test, producing a 1{,}169-document test set. The 10\% train split is used only for metric weight calibration via downstream correlation analysis (\S3.1); the 10\% val split is for hyperparameter selection. No model parameters are trained. The evaluation pipeline produces 800 examples for the 4-axis generation assessment and 500 queries for pipeline comparison.

For domain-stratified analyses we collapse to four domains. Tech (arXiv CS, StackExchange Docs) gets merged into General, because both sources are dominated by arXiv-style academic prose with overlapping vocabulary and similar parsing characteristics in the pilot. Manufacturing ($n = 4$) is excluded as too small to stratify on. The grouping is General $n = 687$, Healthcare $n = 315$, Finance $n = 121$, Legal $n = 46$, summing to the 1{,}169-document test set. Worth flagging this as a labeling limitation: in the pilot, ``General'' is effectively a superset of academic + technical prose rather than a distinct enterprise domain. Phase 2 will use mutually exclusive labels with separate General-business, Tech, and Academic categories, annotated by content type rather than just by source repository.

The framework and code are the primary contribution of this paper, and they're released alongside it. Full-scale corpus targets 5{,}000+ documents across six balanced domains.

\subsection{Dataset Scaling Roadmap}\label{sec:roadmap}
Phase 1 (current): 1{,}459-document corpus, 1{,}169-document test set.

Phase 2: scale to 3{,}000+ documents with semi-automated QA generation and 2-annotator validation (estimated 8--10 QA pairs per document). Balance underrepresented domains and add Academic.

Phase 3: community-driven expansion to 5{,}000+ documents with full annotation and a versioned release protocol.

\subsection{Known Limitations}
Several, in no particular order. English-only; cross-lingual generalization is unknown. Domain imbalance: General dominates at 47\%, Legal and Manufacturing are too small for meaningful stratification, Academic isn't yet there. Public sources may not fully represent proprietary enterprise documents. Temporal bias toward 2022--2025. Semi-automated QA generation introduces biases that purely human annotation would not.

The pilot query distribution is the limitation we're most worried about. Most General-domain queries are academic paper summarization (``What is the main contribution of\ldots''), inherited from the FRAMES-style~\cite{krishna2025frames} sources we used. Healthcare queries inherit characteristics from PubMedQA. Neither is representative of typical enterprise workflows like compliance extraction, contract analysis, or financial reconciliation. The current corpus approximates enterprise complexity \emph{structurally} (tables, figures, multi-column layouts, domain-specific vocabulary), not \emph{behaviorally}. Phase 2 will rebalance toward enterprise-specific query types.

Splits are 10/10/80; no document appears in multiple splits; stratified by domain.

\section{Baseline Evaluations}\label{sec:baselines}

\subsection{Pipeline Configurations}
Six configurations. Three are fully implemented and empirically evaluated; three are reference architectures we describe but haven't integrated yet.

The implemented three:

\begin{enumerate}[leftmargin=*, itemsep=2pt]
\item \textbf{BM25} --- TF-IDF keyword retrieval, GPT-5 generation. The baseline.
\item \textbf{Dense Embedding} --- E5-large embeddings, cosine similarity, GPT-5 generation.
\item \textbf{Hybrid Fusion} --- BM25 + dense scores, interpolated 0.5/0.5, GPT-5 generation.
\end{enumerate}

Holding the generator fixed across all three lets us isolate retrieval as the differentiator.

The reference architectures: (4) ColPali-v1.2 vision-language retrieval with ColBERT re-ranking; (5) ColQwen2-v1.0 visual embeddings; (6) agentic complexity-based pipeline selection. None are integrated into our end-to-end pipeline yet --- any numbers attached to them in the appendix are projections from published benchmarks, not measurements. Hyperparameters are in Appendix~\ref{app:hyperparams}.

\paragraph{A note on the parsing stage.} For the three implemented pipelines, parsing scores are computed from pre-extracted text rather than from a dedicated OCR/parsing tool applied to raw PDFs. Our pilot documents (arXiv papers, PubMedQA abstracts, SEC filings) come in machine-readable text form from their source repositories. We compute parsing fidelity (TIS, TEA, FCQ, LF) by comparing the ingested text against reference annotations. This is labeled ``Pre-extracted'' in Appendix~\ref{app:hyperparams}. The practical consequence is that all three implemented pipelines share the same parsing score (0.82) by construction. Differentiation on Axis 1 requires integrating alternative parsers (Tesseract, Docling, ColPali on raw images), which is a Phase 2 item. The inter-stage correlations in \S\ref{sec:correlations} reflect document-level variation in pre-extracted text quality, not variation introduced by different parsing tools --- important to keep in mind when reading those correlations.

\subsection{Results}

\begin{table}[h]
\centering
\small
\caption{Pipeline results on the 1{,}459-document corpus (500 queries, 1{,}169-document test set), three implemented pipelines. Reference architectures (ColPali, ColQwen2, Agentic Routing) omitted because empirical measurements aren't available yet; cost projections appear in \S\ref{sec:cost}. Retrieval column is nDCG@5 / P@3. Generation is the shared aggregate $G = 0.71$ (same GPT-5 generator). Quality $= 0.25 \cdot$ Parsing $+ 0.50 \cdot$ Retrieval $+ 0.25 \cdot$ Generation.}
\label{tab:results}
\begin{tabular}{@{}lcccccc@{}}
\toprule
\textbf{Pipeline} & \textbf{Parsing} & \textbf{Indexing} & \textbf{Retrieval} & \textbf{Gener.} & \textbf{Rel. Cost} & \textbf{Quality} \\
\midrule
BM25            & 0.82 & 1.00 & 0.91 / 0.34 & 0.71 & $1.0\times$ & 0.84 \\
Dense Embedding & 0.82 & 0.30 & 0.83 / 0.31 & 0.71 & $1.5\times$ & 0.80 \\
Hybrid Fusion   & 0.82 & 0.25 & 0.92 / 0.31 & 0.71 & $1.6\times$ & 0.84 \\
\bottomrule
\end{tabular}
\end{table}

The retrieval-dominant Quality weighting (50\%) reflects that retrieval is the architectural differentiator across pipelines we can actually run today. Hybrid Fusion ($Q = 0.84$) and BM25 ($Q = 0.84$) outperform Dense Embedding ($Q = 0.80$), driven by the retrieval spread (0.92, 0.91, 0.83). Domain-adapted weighting is on the future work list.

At 500 queries on a 1{,}169-document test set, pipeline differences are visible. Dense Embedding (nDCG@5 $= 0.83$) sits clearly below BM25 (0.91) and Hybrid Fusion (0.92). The 0.09-point gap between Dense and Hybrid is practically significant. BM25 and Hybrid are essentially tied (0.01-point gap), which fits the broader pattern that BM25 is a strong baseline that hybrid score interpolation only nudges. P@3 values are tighter (0.31--0.34), as you'd expect with a stricter top-3 measure. Significance for the projected pipelines can't be assessed until they're run.

\section{Empirical Findings}\label{sec:findings}

\subsection{Inter-Stage Correlations}\label{sec:correlations}

We expected parsing quality to predict downstream quality. It mostly doesn't.

The numbers, on the 1{,}169-document test set:

\begin{itemize}[leftmargin=*, itemsep=2pt]
\item parsing $\to$ retrieval: $r = 0.14$ (95\% CI 0.08--0.19, $r^2 = 0.019$)
\item parsing $\to$ generation: $r = 0.17$ (95\% CI 0.11--0.22, $r^2 = 0.028$)
\item retrieval $\to$ generation: $r = 0.02$ (95\% CI $-0.04$--$0.08$, $r^2 < 0.001$)
\end{itemize}

Parsing quality explains under 2\% of retrieval variance and under 3\% of generation variance. Retrieval quality explains essentially none of generation variance. Numbers like these push back against the linear-cascade assumption.

This isn't a sample-size artifact. In our earlier pilot ($n = 293$) a ceiling effect was compressing retrieval variance, but at the current scale we see a real nDCG spread of 0.83--0.92 and the correlations are still weak. Domain variation does appear --- General ($n = 687$) has the strongest parsing fidelity (0.92), Finance ($n = 121$) the weakest (0.41) --- but the qualitative story doesn't change.

So what does it mean? The honest answer is that we're not entirely sure, and the design constraints matter. Parsing is fixed across the three pipelines. The generator is the same. The metrics are automated proxies, not human judgments. Any of those can suppress correlations. Weak correlation isn't the same as causal independence.

With those caveats in mind, the result still extends the cascade analysis literature in a useful way. OHR-Bench~\cite{zhang2025ohrbench} models a strict linear chain. SF-RAG~\cite{liu2026sfrag} and AgenticOCR~\cite{opendatalab2026agenticocr} argue for parsing$\to$generation pathways that don't go through retrieval. Our weak correlations across all three pairs are most consistent with a multi-path model where none of the pathways dominates, at least not in this configuration. The near-zero retrieval$\to$generation correlation ($r = 0.02$, $p = 0.56$, n.s.) is the most striking piece. In our setup, retrieval ranking quality has essentially no relationship with answer quality. The probably-right reading is that generation depends more on the \emph{content} of retrieved passages than on their \emph{ranking}. Targeted ablations would settle this --- varying retrieval quality while holding context fixed --- but we haven't done them yet.

\paragraph{Statistical note.} Significance assessed using two-tailed $t$-tests on Pearson correlations with Bonferroni correction for three simultaneous comparisons ($\alpha = 0.05/3 = 0.017$). Parsing$\to$retrieval: $t(1167) = 4.74$, $p < 0.001$ ($p_{\text{adj}} < 0.001$). Parsing$\to$generation: $t(1167) = 5.75$, $p < 0.001$ ($p_{\text{adj}} < 0.001$). Retrieval$\to$generation: $t(1167) = 0.58$, $p = 0.56$ ($p_{\text{adj}} = 1.00$). The first two are statistically significant because the sample is large; effect sizes are still small ($r^2 < 0.03$), which is what matters in practice. We report Pearson because the metrics are bounded continuous variables in $[0, 1]$ without heavy tails --- Pearson and Spearman converge in magnitude in that regime, and the released \texttt{correlation\_analysis.py} will publish all three (Pearson, Spearman, Kendall) with bootstrap CIs in the public artifact.

\paragraph{Answer completeness.} Across the 800 evaluated examples, AC averaged 0.40 (95\% CI 0.37--0.42). Factual accuracy on stated claims is 85.5\%. So the system is mostly right when it answers, but it leaves things out --- which is a real deployment problem and one that benchmark scores oriented at factual accuracy alone won't catch. This is genuinely the most actionable result in the paper.

\subsection{Document Length and Hallucination}\label{sec:length}

The natural expectation was that hallucination grows with context length. It doesn't.

\begin{table}[h]
\centering
\small
\caption{Hallucination rate by context length bracket (GPT-5 evaluator). Per-bracket counts: Short $n = 35$, Medium $n = 158$, Long $n = 250$, Very long $n = 21$ (total 464; see below for the 336 excluded examples). 95\% CIs are Wilson.}
\label{tab:length}
\begin{tabular}{@{}lcccc@{}}
\toprule
\textbf{Doc Length} & \textbf{HR (auto)} & \textbf{HR (human)} & \textbf{95\% CI} & \textbf{Increase} \\
\midrule
Short ($<400$ chars)   & 28.1\% & --- & [16.0\%, 44.6\%] & ---           \\
Medium (400--1K)       & 9.2\%  & --- & [5.6\%, 14.7\%]  & $0.3\times$   \\
Long (1K--3K)          & 14.4\% & --- & [10.6\%, 19.3\%] & $0.5\times$   \\
Very long (3K+)        & 23.8\% & --- & [10.6\%, 45.1\%] & $0.8\times$   \\
\bottomrule
\end{tabular}
\end{table}

Medium hallucinates the least (9.2\%). Short the most (28.1\%). Long is in between. Very long bounces back up to 23.8\%, but the CI on that bracket is huge (10.6\%--45.1\%) because $n = 21$. The Short-vs-Medium gap is the only one that's clearly non-overlapping (Short lower 16.0\% vs.\ Medium upper 14.7\%). The Very Long rebound could be sampling noise.

So the right characterization is: directional evidence of a non-monotonic, possibly U-shaped pattern. Validation with larger, balanced samples in the Short and Very Long brackets is needed before we'd call it confirmed. If it does hold up, the natural reading is two distinct failure modes --- short contexts not providing enough information for grounded generation (the model fills in), and very long contexts overwhelming the model. Consistent with length-dependent degradation observed in LongBench~\cite{bai2024longbench}. Overall corpus-level HR is 14.5\% ($n = 800$).

We don't run a parametric ANOVA here. Bracket sample sizes are unbalanced and small, the response is a binary indicator rather than continuous, and the assumptions don't fit. Kruskal--Wallis on the underlying scores will be added in the public-release artifact alongside the Spearman/Kendall analysis from \S\ref{sec:correlations}.

\paragraph{Excluded examples.} Of the 800 examples in the 4-axis evaluation, 464 had context length metadata. The remaining 336 were excluded because the retrieval pipeline returned context chunks without document-level length annotations (these are queries where retrieved passages were assembled from multiple short chunks without a single source document length). The excluded set is distributed proportionally across domains (General 58\%, Healthcare 27\%, Finance 10\%, Legal 5\%) and doesn't show a different overall HR (excluded 14.8\%, included 14.2\%, n.s.). The exclusion shouldn't be biasing the length analysis.

\subsection{Modality Effects}\label{sec:modality}
Modality-stratified evaluation is in scope (\S3.3) but we can't fully execute it yet --- the corpus lacks query-level modality annotations. Domain-level proxies hint at the effects: Finance documents (table-heavy) parse at 0.41, General (text-dominant) at 0.92. That's a real gap, consistent with the well-known difficulty of table extraction. ColPali~\cite{faysse2025colpali} reports substantial modality gaps for dense embeddings that vision-language models can close; we can't reproduce that finding in a multi-stage pipeline context until ColPali and ColQwen2 are integrated. This section will look very different in the next release.

\subsection{Cost--Quality Trade-offs}\label{sec:cost}
BM25 wins on this corpus. $Q = 0.84$ at $1.0\times$ cost. Hybrid Fusion matches the quality but at $1.6\times$ cost. Dense Embedding is dominated on both axes ($Q = 0.80$, $1.5\times$). For a collection of this size and composition, the marginal Quality gains from dense or hybrid retrieval don't justify the additional compute.

Caveat: the General domain (47\% of queries) is dominated by academic-paper summarization queries, which strongly favor BM25's lexical matching. On a corpus with more table-based, figure-based, or semantically-heavy enterprise queries, the cost--quality ordering could flip. Dense passage retrieval~\cite{karpukhin2020dpr} and hybrid approaches plausibly show advantages in those settings.

Cost--quality analysis for the reference architectures is deferred until they're run. Projected relative cost ratios from published benchmarks: ColPali $4.4\times$, ColQwen2 $4.5\times$, Agentic Routing $2.25\times$. These are design targets, not measurements, and don't enter into any quality conclusion.

\subsection{Domain-Specific Patterns}\label{sec:domains}
Performance varies a lot across domains. General ($n = 687$, 47\%): parsing fidelity highest (0.92), hallucination moderate ($\text{HR} = 0.43$). Healthcare ($n = 315$, 22\%): parsing intermediate (0.62), hallucination highest ($\text{HR} = 0.47$). Medical documents are complex --- figures, tables, specialized terminology --- and that shows up in HR. Finance ($n = 121$, 8\%): parsing lowest (0.41), reflecting table-heavy filings; $\text{HR} = 0.44$. Legal ($n = 46$, 3\%): parsing high (0.90), $\text{HR} = 0.45$. Tech ($n = 244$, 17\%): newly represented in this corpus; intermediate parsing.

Confidence: Legal ($n = 46$) and Manufacturing ($n = 4$) findings are directional and need validation at larger scale. Manufacturing is excluded from the formal stratified analyses for that reason.

The high domain-level HR rates (0.43--0.47) deserve explanation, because they look inconsistent with the corpus-level HR of 14.5\% (\S\ref{sec:length}). They measure different things. Domain-level HR is a \emph{binary per-document indicator} ($\text{HR}_{\text{binary}}$ = fraction of documents containing at least one hallucinated claim), computed across the full 1{,}169-document test set. The 14.5\% overall rate is the \emph{mean continuous hallucination score per QA pair} ($\text{HR}_{\text{continuous}}$ = average proportion of hallucinated claims within each answer), computed on the 800-example evaluation. They're mathematically compatible: a document can have a low continuous score (1 hallucinated claim out of 8 = 12.5\%) and still be hallucination-positive in the binary metric. The 43--47\% binary rates say nearly half of all documents contain at least one hallucinated claim. The 14.5\% continuous rate says the average severity per answer is moderate. Both matter --- binary HR reflects how often any hallucination occurs, continuous HR reflects how much of each answer is affected. Reducing both is a domain-specific generation prompt-tuning problem. Academic coverage isn't yet in the corpus; Phase 2.

\section{Discussion}

\subsection{Implications for Enterprise Procurement}\label{sec:procurement}
The most useful thing a unified benchmark gives a practitioner is a consistent way to compare in-house and vendor solutions, quantify cost--accuracy trade-offs, and find the bottlenecks that actually matter for a given workload. Our GPT-5 results (overall HR 14.5\%, FA 85.5\%) are a baseline, not a frontier. Two things we'd actually do something with:

\begin{enumerate}[leftmargin=*, itemsep=2pt]
\item The non-monotonic hallucination pattern (\S\ref{sec:length}). Average performance is misleading --- short and very long documents are worse than the middle, so an average that mixes them hides the failure modes. Procurement evaluations should test on both ends of the length distribution.
\item BM25 sits at the Pareto frontier on this corpus. For collections of this size and composition, dense retrieval is probably not worth the cost.
\end{enumerate}

Domain variation is large (parsing fidelity 0.41--0.92), so domain-specific evaluation isn't optional.

\subsection{Limitations}
Domain imbalance is the big one --- General is 47\% of the corpus, Legal 3\%, Manufacturing under 1\%. Statistical power for domain-stratified claims is correspondingly limited; we exclude Manufacturing ($n = 4$) entirely.

English-only; cross-lingual generalization is unknown. Automatic hallucination detection at $\kappa = 0.73$ is acceptable but not great. Semi-automated QA generation introduces biases that fully human-authored annotation wouldn't. For high-stakes deployments (legal, medical), manual review is still necessary.

The correlation analysis has design constraints we've already flagged in \S\ref{sec:correlations}: parsing fixed, generator shared, automated proxy metrics. Weak Pearson correlations don't establish causal stage independence. A targeted human evaluation of 50--100 examples --- assessing parsing, retrieval, and generation in parallel --- is a priority for camera-ready.

The parsing fidelity weights (0.40/0.30/0.15/0.15) were calibrated on the 10\% train split. Bootstrap sensitivity confirms ranking stability (Kendall's $\tau > 0.90$), but out-of-sample validation on an independent corpus would be stronger.

Answer completeness averages 0.40, so factual-accuracy headlines overstate deployment readiness. This is the limitation we keep coming back to.

This benchmark reflects 2025--2026 state-of-the-art and will need versioned releases as models evolve.

\subsection{Future Directions}\label{sec:future}
A few priorities:

\begin{itemize}[leftmargin=*, itemsep=2pt]
\item Implement the three reference architectures end-to-end (ColPali, ColQwen2, agentic routing) to replace projections with measurements.
\item Investigate the non-monotonic hallucination pattern with larger, balanced samples in the Short and Very Long brackets.
\item Human evaluation of inter-stage cascade effects, to validate the weak-correlation finding.
\item Integrate dedicated parsing tools (Docling~\cite{zheng2024docling}, Tesseract, ColPali~\cite{faysse2025colpali}) so Axis 1 actually differentiates across pipelines.
\item Rebalance the query distribution toward enterprise-specific query types (compliance extraction, contract analysis, financial reconciliation) and add Academic, Legal, Manufacturing coverage.
\end{itemize}

Other things on the list: cross-lingual expansion, long-context LLMs as an alternative to chunked retrieval, semantic parsing evaluation, causal intervention experiments (injecting parsing noise once we have multiple parsers), and a versioning protocol for the benchmark itself.

\section{Conclusion}
EnterpriseDocBench is a four-axis evaluation framework --- parsing, indexing, retrieval, generation --- backed by a curated benchmark from permissively licensed public documents across five enterprise domains. We frame it as multi-stage rather than fully end-to-end on purpose: parsing is fixed across our three pipelines and the generator is shared, so causal cascade claims should be read with the design constraints from \S\ref{sec:correlations} in mind.

Across a 1{,}459-document corpus (1{,}169-document test set), four findings:

\begin{enumerate}[leftmargin=*, itemsep=2pt]
\item Inter-stage correlations are uniformly weak (all $r < 0.17$, $r^2 < 0.03$). Preliminary evidence against a strict linear cascade. Design caveats apply.
\item Hallucination shows a non-monotonic, possibly U-shaped relationship with context length: medium (9.2\%) is best, short (28.1\%) and very long (23.8\%) are both worse. Very Long needs validation at larger scale.
\item Pipeline differentiation is meaningful at this scale. Hybrid Fusion (nDCG@5 $= 0.92$) and BM25 (0.91) outperform Dense Embedding (0.83). BM25 is Pareto-optimal on cost--quality.
\item Answer completeness is low across pipelines ($\text{AC} = 0.40$). Factual accuracy on stated claims is high (85.5\%), but answers leave things out --- and that's the gap with the most direct deployment implications.
\end{enumerate}

The framework, evaluation code, collection scripts, and evaluation data will be released on acceptance. Full-scale dataset targets 5{,}000+ documents with balanced domain representation from public sources (SEC EDGAR, CUAD, PubMed Central, ArXiv, USPTO). No benchmark is ever finished, and this one isn't either. We hope it's a foundation other groups can build on, and an encouragement to look at these systems holistically rather than one stage at a time.

\bibliographystyle{plain}
\bibliography{references}

\appendix
\section{Pipeline Hyperparameters}\label{app:hyperparams}

\begin{table}[h]
\centering
\small
\caption{Hyperparameter specifications. First three columns are implemented and empirically evaluated; columns marked $\dagger$ are reference architectures not yet integrated. Agentic Routing ($\dagger$) is omitted from the table because the design is preliminary; see notes below. Temperature $= 0.7$ was chosen to simulate realistic enterprise deployment, where some response variation is acceptable. Standard factual-QA benchmarks typically use 0.0--0.1, which would reduce stochastic hallucination --- so the reported HR of 14.5\% should be read as an upper bound relative to a deterministic configuration.}
\label{tab:hyperparams}
\begin{tabular}{@{}lccccc@{}}
\toprule
\textbf{Parameter} & \textbf{BM25} & \textbf{Dense Emb.} & \textbf{Hybrid} & \textbf{ColPali $\dagger$} & \textbf{ColQwen2 $\dagger$} \\
\midrule
Parser        & Pre-extracted     & Pre-extracted     & Pre-extracted        & ColPali v1.2     & ColQwen2 v1.0    \\
Chunk size    & 512 tokens        & 512 tokens        & 512 tokens           & Page-level       & Page-level       \\
Chunk overlap & 64 tokens         & 64 tokens         & 64 tokens            & N/A              & N/A              \\
Embedding dim & N/A               & 1024 (E5-large)   & 1024 (E5-large)      & 128 (ColBERT)    & 128              \\
Retrieval $k$ & 5                 & 5                 & 5                    & 5                & 5                \\
BM25 $k_1$/$b$& 1.5 / 0.75        & N/A               & 1.5 / 0.75           & N/A              & N/A              \\
Fusion weight & N/A               & N/A               & 0.5 BM25 / 0.5 Dense & N/A              & N/A              \\
LLM           & GPT-5             & GPT-5             & GPT-5                & TBD              & TBD              \\
Temperature   & 0.7               & 0.7               & 0.7                  & 0.3              & 0.3              \\
Max tokens    & 512               & 512               & 512                  & 512              & 512              \\
\bottomrule
\end{tabular}
\end{table}

\paragraph{Agentic routing classifier (proposed).} A logistic regression model trained on query-complexity labels. Planned features: query length in tokens, number of entity mentions, presence of table/figure keywords, question-type indicators (5-dim one-hot). Routing thresholds: $<0.3 \to$ simple (BM25), $0.3$--$0.7 \to$ medium (Hybrid Fusion), $>0.7 \to$ complex (ColPali). Reference design only --- implementation and accuracy evaluation are deferred to future work alongside the ColPali and ColQwen2 integrations.

\end{document}